\documentclass[runningheads]{llncs}
\usepackage{graphicx}
\usepackage{amsmath}
\usepackage{booktabs}
\usepackage{xcolor}
\usepackage[inline]{enumitem}

\usepackage{multirow}

\usepackage{comment}

\newcommand{\todo}[1]{{\leavevmode\color{red}#1}}

\newcommand{\remove}[1]{}

\usepackage[numbers,sort&compress]{natbib}

\begin{document}
\title{Revisiting Non-Specific Syndromic Surveillance}
%
%
\author{Moritz Kulessa\inst{1}
\and Eneldo Loza Menc\'{i}a\inst{1}
\and Johannes F\"urnkranz\inst{2}
}
\authorrunning{M. Kulessa et al.}
%
\institute{Technische Universit\"at Darmstadt, Germany \email{\{mkulessa,eneldo\}@ke.tu-darmstadt.de} \and
Johannes Kepler Universit\"at Linz, Austria \\
\email{juffi@faw.jku.at}}
\maketitle              
\begin{abstract}
Infectious disease surveillance is of great importance for the prevention of major outbreaks. Syndromic surveillance aims at developing algorithms which can detect outbreaks as early as possible by monitoring data sources which allow to capture the occurrences of a certain disease. Recent research mainly focuses on the surveillance of specific, known diseases, putting the focus on the definition of the disease pattern under surveillance. Until now, only little effort has been devoted to what we call \emph{non-specific} syndromic surveillance, i.e., the use of all available data for detecting any kind of outbreaks, including infectious diseases which are unknown beforehand. In this work, we revisit published approaches for non-specific syndromic surveillance and present a set of simple statistical modeling techniques which can serve as benchmarks for more elaborate machine learning approaches. Our experimental comparison on established synthetic data and real data in which we injected synthetic outbreaks shows that these benchmarks already achieve very competitive results and often outperform more elaborate algorithms.
\end{abstract}

\section{Introduction}

The early detection of infectious disease outbreaks is of great significance for public health. The spread of such outbreaks could be diminished tremendously by applying control measures as early as possible, which indeed can save lives and reduce suffering. 
For that purpose, \emph{syndromic surveillance} has been introduced which aims to identify illness clusters before diagnoses are confirmed and reported to public health agencies~\cite{ss_henning}. 

The fundamental concept of syndromic surveillance is to define indicators for a particular infectious disease on the given data, also referred to as \emph{syndromes}, which are monitored over time to be able to detect unexpectedly high numbers of infections which might indicate an outbreak of that disease. Syndromic data can be obtained from \emph{clinical} data sources (e.g., diagnosis in an emergency department), which allow to directly measure the symptoms of individuals, as well as \emph{alternative} data sources (e.g., internet-based health inquiries), which indirectly capture the presence of a disease~\cite{ss_henning}. 

In general, the definition of syndromes is a challenging task since symptoms are often shared by different diseases and a particular disease can have different disease patterns in the early phase of an infection. Moreover, this kind of filtering is a highly handcrafted approach and only allows to monitor known infectious diseases. Rather than developing highly specialized algorithms which are based on a specific disease and assume particular characteristics of outbreak shapes~\cite{ss_shemueli}, we argue that the task of outbreak detection should be viewed as a general anomaly detection problem where an outbreak alarm is triggered if the distribution of the incoming data changes in an unforeseen and unexpected way. Therefore, we distinguish between \emph{specific} syndromic surveillance, where factors related to a specific disease are monitored, and \emph{non-specific} syndromic surveillance, where general, universal characteristics of the stream of data are monitored for anomalies. While specific syndromic surveillance is a well-studied research area, we found that only little research has been devoted to non-specific syndromic surveillance with only very few algorithms available. In particular, the close relation to anomaly detection motivated us to investigate the problem of non-specific syndromic surveillance from a machine learning perspective and to make the task more approachable for the anomaly detection community.

In this paper, we revisit algorithms for non-specific syndromic surveillance and compare them to a broad range of anomaly detection algorithms. Due to little effort on implementing baselines in previous works on non-specific syndromic surveillance, we propose a set of benchmarks relying on simple statistical assumptions which nonetheless have been widely used before in specific syndromic surveillance. 
We experimentally compare the methods on an established synthetic dataset~\cite{nss_fanaee, nss_wong3} and real data from a German emergency department in which we injected synthetic outbreaks.
Our results demonstrate that the simple statistical approaches, which have not been considered in previous works, are quite effective and often can outperform more elaborate machine learning algorithms.
\section{Non-Specific Syndromic Surveillance}


\subsection{Problem Definition}
\label{sec:problem_definition}
Syndromic data can be seen as a constant stream of instances of a population $\mathcal{C}$. Each instance $\mathbf{c} \in \mathcal{C}$ is represented by a set of attributes $\mathcal{A} = \{A_1, A_2, \ldots, A_m\}$ where each attribute can be either categorical (e.g., \textsl{gender}), continuous (e.g., \textsl{age}) or text (e.g., \textsl{chief complaint}). Following the notation of~\citet{nss_wong3}, we refer to these attributes as \emph{response attributes}. To be able to detect changes over time, instances are grouped together according to pre-specified time slots (e.g., all patients arriving at the emergency department in one day). Hence, the instances for a specific time slot $t$ are denoted as $\mathcal{C}(t) \subset \mathcal{C}$. 

In addition, each group $\mathcal{C}(t)$ is associated with an environmental setting $\mathbf{e}(t) \in E_1 \times E_2 \times \ldots \times E_k$ where $\mathcal{E} = \{E_1, E_2, \ldots, E_k\}$ is a set of \emph{environmental attributes}. Environmental attributes are independent of the response attributes and represent external factors which might have an influence on the distribution of instances $\mathcal{C}(t)$ (e.g., during the winter flu-like symptoms are more frequent). In particular, a specific characteristic of syndromic data is \emph{seasonality}, in machine learning also known as \emph{cyclic drift}~\citep{ml_ra_webb}. Environmental variables can help the algorithm to adapt to this kind of concept drift. Thus, the information available for time slot $t$ can be represented by the tuple $(\mathcal{C}(t), \mathbf{e}(t))$ and the information about prior time slots can be denoted as $\mathcal{H} = ((\mathcal{C}(1), \mathbf{e}(1)), \ldots, (\mathcal{C}(t-1), \mathbf{e}(t-1)))$. 

The main goal of non-specific syndromic surveillance is to detect 
anomalies in the set $\mathcal{C}(t)$ of the current time slot $t$ w.r.t.\ the previous time slots $\mathcal{H}$ as potential indicators of 
an infectious disease outbreak. Therefore, the history $\mathcal{H}$ is used to fit a model $f_{\mathcal{H}}(\mathbf{e}(t), \mathcal{C}(t))$ which is able to generate a score for time slot $t$, representing the likelihood of being in an outbreak. 

Viewed from the perspective of specific syndromic surveillance, the non-specific setting can be seen as the monitoring of all possible syndromes at the same time. The set of all possible syndromes can be defined as 
\begin{align*}
    \mathcal{S} = 
    \left\{
    \prod_{i \in \mathcal{I}} A_i \mid A_i \in \mathcal{A} \wedge \mathcal{I} \subseteq \{1,2, \ldots, m\} \wedge |\mathcal{I}| \geq 1
    \right\}
\end{align*}
where $\prod_{i \in\mathcal{I}} A_i$ for $|\mathcal{I}| = 1$ is defined as $\{\{a\} \mid a \in A \wedge A \in \mathcal{A}\}$. In addition, we denote $\mathcal{S}_{\leq n} = \{s \mid s \in \mathcal{S} \wedge |s| \leq 2 \}$ as the set of all possible syndromes having a maximum of $n$ conditions and $\mathcal{H}_s=(s(1), s(2), \ldots, s(t-1))$ as the time series of counts for a particular syndrome $s \in \mathcal{S}$. 

\subsection{Evaluation}
\label{sec:evaluation}

To evaluate a data stream it is split into two parts, namely a \emph{training part}, containing the first time slots which are only used for training, and a \emph{test part}, which contains the remaining time slots of the data stream. The evaluation is performed on the test part incrementally which means that for evaluating each time slot $t$ the model will be newly fitted on the complete set of previously observed data points $\mathcal{H} = ((\mathcal{C}(1), \mathbf{e}(1)), \ldots, (\mathcal{C}(t-1), \mathbf{e}(t-1)))$. Alarms raised during an outbreak are considered as true positives while all other raised alarms are considered as false positives.

For measuring the performance, we rely on the \emph{activity monitor operating characteristic (AMOC)}~\cite{ss_e_fawcett}. AMOC can be seen as an adaptation of the \emph{receiver operating characteristic} in which the true positive rate is replaced by the \emph{detection delay}, i.e., the number of time points until an outbreak has been first detected by the algorithm. In case the algorithm does not raise an alarm during the period of an outbreak, the detection delay is equal to the length of the outbreak. 
Moreover, for syndromic surveillance we are interested in a very low false alarm rate for the algorithms and therefore only report the partial area under AMOC-curve for a false alarm rate less than $5\%$, to which we refer to as $AAUC_{5\%}$. Note that contrary to conventional AUC values in this case lower values represent better results. Since one data stream does normally not contain enough outbreaks to draw conclusions, the evaluation is usually performed on a set of data streams. To obtain a final score for the set, we take the average over the computed $AAUC_{5\%}$ results which are computed on each data stream.

\section{Machine Learning Algorithms}
\label{sec:approaches}


In a survey of the relevant literature we have identified only a few algorithms which relate to non-specific syndromic surveillance, described in Sections \ref{sec:dmss} to \ref{sec:eigenevent}. In Section \ref{sec:anomaly_detection_algorithms} we introduce a way how common anomaly detection algorithms can be applied in the setting of non-specific syndromic surveillance.

\subsection{Data Mining Surveillance System (DMSS)}
\label{sec:dmss}

One of the first algorithms 
able to identify new and interesting patterns in syndromic data was proposed by \citet{nss_brossette} who adopted the idea of association rule mining \cite{AssociationRules-Book}
to the field of public health surveillance.
In order to detect an outbreak for time slot $t$, an association rule mining algorithm needs to be run on $\mathcal{C}(t)$ and a reference set of patients $\mathcal{R} \subset \mathcal{C}$ is created by merging the instances of a selected set of previous time slots. For each association rule the confidence of the rule on $\mathcal{C}(t)$ is compared to the confidence of the rule computed on $\mathcal{R}$ using a $\chi^2$ or a Fisher's test. If the confidence has significantly increased on $\mathcal{C}(t)$, the finding is reported as an unexpected event. In order to reduce the complexity, the authors propose to focus only on mining high-support association rules. An aggregation of the observations for one time slot is not performed and environmental attributes are not considered by this approach.

\subsection{What is strange about recent events? (WSARE)}
\label{sec:wsare}

The family of \emph{WSARE} algorithms has been proposed by \citet{nss_wong3}. All algorithms share the same underlying concept, namely to monitor all possible syndromes having a maximum of two conditions $\mathcal{S}_{\leq 2}$ simultaneously. The three WSARE algorithms only differ in the way how the reference set of patients $\mathcal{R}$ is created on which the expected proportion for each syndrome is estimated. Each expected proportion is compared to the proportion of the respective syndrome observed on the set $\mathcal{C}(t)$ using the $\chi^2$ or Fisher's exact test. In order to aggregate the $p$-values of the statistical tests for one time slot, a \emph{permutation test} with 1,000 repetitions is performed.
The following three versions have been considered:
\begin{description}
\item[WSARE 2.0] merges the instances of a selected set of prior time slots together for the reference set $\mathcal{R}$. Since their evaluation was based on single-day time slots, they combined the instances of the previous time slots $35$, $42$, $49$ and $56$ to consider only instances of the same weekday.
\item[WSARE 2.5] merges the instances of all prior time slots together which share the same environmental setting as for the current day $\mathbf{e}(t)$. This has the advantage that the expected proportions are conditioned on the environmental setting $\mathbf{e}(t)$ and that potentially more instances are contained in the reference set $\mathcal{R}$, allowing to have more precise expectations. 
\item[WSARE 3.0] learns a Bayesian network over all recent data $\mathcal{H}$ from which 10,000 instances for the reference set $\mathcal{R}$ are sampled given the environmental attributes $\mathbf{e}(t)$ as evidence. 
\end{description}

\subsection{Eigenevent}
\label{sec:eigenevent}

The key idea of the \emph{Eigenevent} algorithm proposed by \citet{nss_fanaee} is to track changes in the data correlation structure using eigenspace techniques. Instead of monitoring all possible syndromes, only overall changes and dimension-level changes are observed by the algorithm. Therefore, a dynamic baseline tensor is created using the information of prior time slots $\mathcal{H}$ which share the same environmental setting $\mathbf{e}(t)$. In the next step, information of the instances $\mathcal{C}(t)$ and the baseline tensor are decomposed to a lower-rank subspace in which the eigenvectors and eigenvalues are compared to each other, respectively. Any significant changes in the eigenvectors and eigenvalues between the baseline tensor and the information of instances $\mathcal{C}(t)$ indicate an outbreak.

\subsection{Anomaly Detection Algorithms}
\label{sec:anomaly_detection_algorithms}

A direct application of point anomaly detection is in general not suitable for syndromic surveillance~\citep{nss_wong3} because these methods aim to identify single instances $\mathbf{c} \in \mathcal{C}$ as outliers and could thus, e.g., be triggered by a patient who is over a hundred years old. In order to still apply point anomaly detectors to discover outbreaks, we form a dataset $\mathcal{D}$ using the syndromes $s \in \mathcal{S}$ as features and the respective syndrome counts $\mathcal{H}_s$ as values. Hence, each instance represents the occurrence counts of all syndromes for one particular time slot and the dataset contains $t-1$ instances in total. This dataset can be used to fit an anomaly detector which can be then applied to the instance of syndrome counts for time slot $t$. Hence, an outbreak could be identified by an unusual combination of syndrome counts. In this work, we consider the following anomaly detection algorithms. Due to space restrictions, we refer to \citet{ml_ra_chandola} and \citet{code_pyod} and the references therein for a comprehensive review of the methods.

\begin{description}
\item[One-Class SVM] extends the support vector machine algorithm to perform outlier detection by separating instances $\mathcal{D}$ from the complement of $\mathcal{D}$.

\item[Local Outlier Factor] computes the outlier score for an instance based on how isolated the instance is with respect to the surrounding neighborhood.

\item[Gaussian Mixture Models] approximate the distribution of the dataset $\mathcal{D}$ using a mixture of Gaussian distributions. The outlier score is based on how dense the region of the evaluated instance is.

\item[Copula-Based Outlier Detection](COPOD) creates an empirical copula for the multi-variate distribution of $\mathcal{D}$ on which tail probabilities for an instance can be predicted to estimate the outlier score. 

\item[Isolation Forest] constructs an ensemble of randomly generated decision trees in which anomalies can be identified by counting the number of splittings required to isolate an instance. 

\item[Autoencoder] learns an identity function of the data through a network of multiple hidden layers. Instances which have a high reconstruction error are considered to be anomalous. 

\item[Multiple-Objective Generative Adversarial Active Learning]~(GAAL)\\ constructs multiple generators having different objectives to generate outliers for learning a discriminator which can assign outlier scores to new instances.

\end{description}

\section{Basic Statistical Approaches}
\label{sec:benchmarks}

In addition to the machine learning models introduced in Section~\ref{sec:approaches}, we also include statistical techniques, which are commonly used for specific syndromic surveillance, into our comparison and adapt them to a non-specific syndromic surveillance setting. 
The key idea of these adaptations is to monitor all possible syndromes $\mathcal{S}$ simultaneously. 
For the purpose of monitoring syndromes, a parametric distribution $P_s(x)$ is fitted for each single syndrome $s \in \mathcal{S}$ using the empirical mean $\mu$ and the empirical variance $\sigma^2$ computed over $\mathcal{H}_s$:
\begin{align*}
\mu = \frac{1}{|\mathcal{H}_s|} \sum_{i=0}^{|\mathcal{H}_s|} s(i) &&
\sigma^2 = \frac{1}{|\mathcal{H}_s|-1} \sum_{i=0}^{|\mathcal{H}_s|} (s(i) - \mu )^2
\end{align*}
On the fitted distribution $P_s(x)$, a one-tailed significance test is performed in order to identify a suspicious increase of cases. For a particular observed count $s(t)$, the $p$-value is computed as the probability $\int_{s(t)}^{\infty} P_s (x) dx$ of observing $s(t)$ or higher counts. Thus, for evaluating a single time slot $t$, we obtain $|\mathcal{S}|$ $p$-values which need to be aggregated under consideration of the multiple-testing problem.
Following \citet{ss_ms_roure}, we only report the minimum $p$-value for each time slot $t$ because the Bonferroni correction can be regarded as a form of aggregation of $p$-values based on the minimum function. 
In particular, note that scale-free anomaly scores are sufficient for the purpose of identifying the most suspicious time slots. 
The complement of the selected $p$-value represents the anomaly score reported for time slot $t$. For our benchmarks we have considered the following distributions:
\begin{description}

\item[Gaussian.] Not tailored for count data but often used in syndromic surveillance is the Gaussian distribution $N(\mu, \sigma^2)$. This distribution will serve as reference for the other distributions which are specifically designed for count data.

\item[Poisson.] The Poisson distribution $Pois(\lambda)$ is directly designed for count data. For estimating the parameter $\lambda$, we use the maximum likelihood estimate which is the mean $\mu$.  

\item[Negative Binomial.] To be able to adapt to overdispersion, we include the negative binomial distribution $NB(r,p)$. We have estimated the parameters with $r=\frac{\mu^2}{\sigma^2-\mu}$ and $p=\frac{r}{r+\mu}$. 

\end{description}
Our preliminary experiments showed that statistical tests on rare syndromes are often too sensitive to changes, causing many false alarms. In addition, outbreaks are usually associated with a high number of infections. Therefore, we set the standard deviation $\sigma^2$ to a minimum of one before fitting the Gaussian distribution and for the Poisson and the negative binomial distribution we set the mean $\mu$ to a minimum of one. We leave the standard deviation untouched for the negative binomial distribution since manipulating the overdispersion can lead to extreme distortions in the estimation.

\section{Experiments and Results}

The goal of the experimental evaluation reported in this section is to provide an overview of the performance of non-specific syndromic surveillance methods in general, and in particular, to re-evaluate the established methods in context of the proposed base statistical approaches and the anomaly detection algorithms. We conducted experiments on synthetic data, which already have been used for the evaluation of the algorithms Eigenevent and WSARE~\citep{nss_fanaee,nss_wong3}, and on real data of a German emergency department (cf.\ Section~\ref{sec:results}). As the emergency department data do not contain any information about real outbreaks, we decided to inject synthetic outbreaks which is common practice in the area of syndromic surveillance, allowing us to evaluate and compare the algorithms in a controlled environment.

\subsection{Evaluation Setup}

\begin{table}[!t]
    \begin{minipage}{.46\linewidth}
      \caption{Information about the \\attributes of the synthetic data.}
      \label{tab:synthetic_data}
    \begin{tabular}{l|l|r}
         attribute & type  & \#values\\
         \hline
         age & response  &  3\\
         gender & response  & 2\\
         action & response &  3\\
         symptom& response & 4\\
         drug & response   & 4\\
         location & response  & 9\\
         \hline
         flu level & environmental & 4\\
         day of week & environmental  & 3\\
         weather & environmental & 2\\
         season & environmental  & 4\\
    \end{tabular}
    \end{minipage}\hfill
    \begin{minipage}{.5\linewidth}
      \caption{Information about the attributes of the real data.}
      \label{tab:real_data}
      \vspace{0.065cm}
    \begin{tabular}{l|l|r}
         attribute & type  &  \#values\\
         \hline
         age & response  &  3\\
         gender & response  & 2\\
         mts & response &  28\\
         fever & response & 2\\
         pulse & response   & 3\\
         respiration & response  & 3\\
         oxygen saturation & response & 2\\
         blood pressure & response  & 2\\
         \hline
         day of week & environmental & 7\\
         season & environmental  & 4\\
    \end{tabular}
    \end{minipage} 
\end{table}

\remove{

\begin{table*}[!t]
    \centering
    \caption{Information about the attributes of the synthetic data.
    }
    \begin{tabular}{lllr}
         \toprule
         attribute & type  & values & cardinality\\
         \midrule
         age & response  & child, senior, working & 3\\
         gender & response  & female, male & 2\\
         action & response & absent, evisit, purchase & 3\\
         symptom& response  & none, nausea, rash, respiratory & 4\\
         drug & response  & none, aspirin, nyquil, vomit-b-gone & 4\\
         location & response  & center, east, west, north, \ldots  & 9\\
         \midrule
         flu level & environmental  & decline, high, low, none & 4\\
         day of week & environmental  & Saturday, Sunday, weekday & 3\\
         weather & environmental & cold, hot & 2\\
         season & environmental  & fall, spring, summer, winter & 4\\
         \bottomrule
    \end{tabular}
    \label{tab:synthetic_data}
\end{table*}

}

\paragraph{Synthetic Data. }
The synthetic data consists of $100$ data streams, generated with the synthetic data generator proposed by~\citet{nss_wong3}. The data generator is based on a Bayesian network and simulates a population of people living in a city of whom only a subset are reported to the data stream at each simulated time slot. Detailed information about the attributes in the data stream is given in Table~\ref{tab:synthetic_data}. Each data stream captures the information about the people on a daily basis over a time period of two years, i.e., each time slot $\mathcal{C}(t)$ contains the patients of one day. In average $34$ instances are reported per time slot and $275$ possible syndromes are contained in the set $\mathcal{S}_{\leq 2}$. The first year is used for the training part while the second year serves as the test part. Exactly one outbreak is simulated in the test part which starts at a randomly chosen day and always lasts for $14$ days. During the outbreak period, the simulated people have a higher chance of catching a particular disease. 



\paragraph{Real Data. } 
We rely on routinely collected, fully anonymized patient data of a German emergency department, captured on a daily basis over a time period of two years. We have extracted a set of response attributes and added two environmental attributes (cf.\ Table~\ref{tab:real_data}). Continuous attributes, such as \emph{respiration}, have been discretized with the help of a physician into meaningful categories. 
In addition, we include the Manchester-Triage-System (MTS)~\citep{ss_ed_graeff} initial assessment which is filled out for every patient on arrival. To reduce the number of values for the attribute MTS, we group classifications which do not relate to any infectious disease, such as various kinds of injuries, into a single value. In average $165$ patients are reported per day and in total $574$ syndromes can be formed for the set $|\mathcal{S}_{\leq 2}|$. 
In preparation for the injection of simulated outbreaks, we replicated the data stream 100 times. For each data stream, we used the first year as the training part and the second year as the test part in which we injected exactly one outbreak. In order to simulate an outbreak, we first uniformly sampled a syndrome from $\mathcal{S}_{\leq 2}$. In a second step, we sampled the size of the outbreak from a Poisson distribution with mean equal to the standard deviation of the daily patient visits and randomly selected the corresponding number of patients from 
all patients that exhibit the sampled syndrome. To avoid over-representing outbreaks on rare syndromes, only $20$ data streams contain outbreaks with syndromes that have a lower frequency than one per day. 
In total, $29$ outbreaks are based on syndromes with one condition and $71$ with two.

\paragraph{Additional Benchmarks.} We also include the \emph{control chart}, the \emph{moving average} and the \emph{linear regression} algorithms into our analysis. Compared to our \emph{syndrome-based} statistical benchmarks, these \emph{global} statistical benchmarks only monitor the total number of instances per time slot and therefore can only give a very broad assessment of outbreak detection performance. For a detailed explanation of these algorithms, we refer to \citet{nss_wong3}.





\begin{table}[t!]
    \centering
    \caption{Results for the $AAUC_{5\%}$ measure on the synthetic data.}
    \label{tab:replicate}
    \begin{tabular}{l||c|c|c|c}
name    &  rerun &   min. $p$-value & permutation test & imported $p$-values \\
\hline
Eigenevent &    4.993 &     -- &        --   &   4.391\\
WSARE 2.0 &     -- &     2.963 &    3.805   &   4.925\\
WSARE 2.5 &     -- &     1.321 &    1.614   &   1.931\\
WSARE 3.0 &     -- &     0.899 &    1.325   &   1.610\\
\end{tabular}
\end{table}

\paragraph{Implementation and Parameterization.}
For the Eigenevent algorithm we rely on the code provided by the authors.\footnote{\url{https://github.com/fanaee/EigenEvent}} All other algorithms are implemented in Python.\footnote{Our code is publicly available at \url{https://github.com/MoritzKulessa/NSS}} Parameters for the DMSS and the anomaly detection algorithms have been tuned in a grid search using $1000$ iterations of \emph{Bootstrap Bias Corrected Cross-Validation}~\cite{ml_ra_tsamardinos} which allows to integrate hyperparameter tuning and reliable performance estimation into a single evaluation loop. The evaluated parameter combinations can be found in our repository. The WSARE, the Eigenevent, the COPOD and the statistical algorithms do not contain any parameters which need to be tuned.

\subsection{Preliminary Evaluation}

In a first experiment, we replicated the experiments on the synthetic data of~\citep{nss_fanaee}. More specifically, we imported and re-evaluated the outlier scores for the synthetic data from the Eigenevent repository (\emph{imported $p$-values}) and compare these to our own results with rerunning the Eigenevent algorithm (\emph{rerun}) and to our implementation of the WSARE algorithms. For the latter, we additionally evaluate the results of just reporting the minimal $p$-value for each time slot (\emph{min.\ $p$-value}, cf.\ Section~\ref{sec:benchmarks}) instead of performing an originally proposed permutation test with $1000$ repetitions (\emph{permutation test}). The results are shown in Table~\ref{tab:replicate}.

Our rerun of the Eigenevent algorithm returned slightly worse results than the imported $p$-values, which could be caused by the random initialization. For the WSARE algorithms, we can observe that our implementation achieves better results than the imported $p$-values, probably due to the different Bayesian network used. In particular, the results for the  minimal $p$-value were better than those for the  more expensive permutation test. Thus, we chose to only report the minimal $p$-value for the WSARE algorithms in the following experiments.



\subsection{Results}
\label{sec:results}

\begin{table}[b!]
    \centering
    \caption{Results for the $AAUC_{5\%}$ measure on the synthetic and real data.}
    \label{tab:results}
    \begin{tabular}{l|l||c|c|c||c|c|c}
           \multirow{2}{*}{category} & \multirow{2}{*}{algorithm name} &     \multicolumn{3}{c||}{synthetic data} & \multicolumn{3}{c}{real data} \\
                           &&   none &    $\mathcal{S}_{\leq1}$ & $\mathcal{S}_{\leq2}$ & none &  $\mathcal{S}_{\leq1}$ &   $\mathcal{S}_{\leq2}$ \\
\hline
\multirow[c]{5}{24mm}{non-specific syndromic surveillance}
&WSARE 2.0 &      -- & 3.028 &  2.963 &      -- & 0.661 & 0.590  \\
&WSARE 2.5 &      -- & 1.099 &  1.321 &      -- & 0.917 & 0.867  \\
&WSARE 3.0 &      -- & 0.803 &  0.899 &      -- & 0.882 & 0.847  \\
&DMSS    &  2.430 &     -- &      -- &  0.953 &     -- &      -- \\
&Eigenevent &  4.993 &     -- &      -- &  0.878 &  -- &     -- \\
\hline              
\multirow{7}{24mm}{anomaly detectors}
&one-class SVM           &      -- & 1.043 &  1.262 &      -- & 0.468 & 0.495  \\
&local outlier factor    &      -- & 2.000 &  2.260 &      -- & 0.642 & 0.610  \\
&Gaussian mixture model  &      -- & 1.117 &  3.547 &      -- & 0.444 & 0.791\\
&isolation forest        &      -- & 4.576 &  4.948 &      -- & 0.873 & 0.835  \\
&COPOD  &      -- & 5.216 &  5.032 &      -- & 0.816 & 0.800  \\
&autoencoder            &      -- & 1.521 &  1.643 &      -- & 0.550 & 0.576  \\
&GAAL                 &      -- & 7.024 & 6.766  &      -- & 0.792 & 0.866  \\
\hline 
\multirow{3}{24mm}{global benchmarks}
&control chart &  5.086 &     -- &      -- &  0.891 &     -- &      -- \\
&moving average &  7.012 &     -- &      -- &  0.910 &     -- &      -- \\
&linear regression &  3.279 &     -- &      -- &  0.819 &     -- &      -- \\
\hline
\multirow{3}{24mm}{syndrome-based benchmarks}
&Gaussian     &      -- & 0.806 &  0.941 &      -- & 0.328 &  0.267 \\
&Poisson      &      -- & 1.294 &  1.347 &      -- & 0.598 &  0.486 \\
&negative binomial  &      -- & 0.895 &  0.958 &      -- & 0.299 &  0.216 \\
\end{tabular}
\end{table}

The results on the synthetic and real data are both shown in Table~\ref{tab:results}. For syndrome-based algorithms, the results for monitoring $\mathcal{S}_{\leq 1}$ and $\mathcal{S}_{\leq 2}$ are reported in the respective columns while results for the other methods are reported in the columns \emph{none}. Note that the worst possible result on the synthetic data is $14$ while for the real data the worst result is $1$. In the first paragraphs, we will discuss the results 
without specifically considering the size of the syndrome sets unless needed. 
The effect of using $\mathcal{S}_{\leq 1}$ or $\mathcal{S}_{\leq 2}$ is discussed in the last paragraph.

\paragraph{Comparison between Non-Specific Syndromic Surveillance Algorithms.} 
Firstly, we analyze the results of the non-specific syndromic surveillance approaches which have been presented in Section~\ref{sec:dmss} to~\ref{sec:eigenevent}. In general, the WSARE algorithms outperform the other algorithms in the group. In particular, the results of the modified versions WSARE 2.5 and WSARE 3.0 on the synthetic data show that the use of environmental attributes can be beneficial. However, the results on the real data indicate the opposite. We further investigated this finding by rerunning WSARE 3.0 on the real data without the use of environmental variables and observed a substantial improvement of the results to $0.613$ for $\mathcal{S}_{\leq1}$ and $0.570$ for $\mathcal{S}_{\leq2}$, respectively. Therefore, we conclude that the modelling of the environmental factors should be done with care since it can easily lead to worse estimates if the real distribution does not follow the categorization imposed by defined attributes. 

The results of the DMSS algorithm suggest that monitoring association rules is not as effective as monitoring syndromes. In particular, the space of possible association rules is much greater than the space of possible syndromes $\mathcal{S}$ which worsens the problem of multiple testing. Especially on the real data this results in a bad performance since the high number of instances per time slot yields too many rules. Conversely, by monitoring only rules with very high support most of the outbreaks remain undetected since the disease pattern could not be captured anymore. 
In contrast to the results reported by~\citet{nss_fanaee}, the Eigenevent algorithm performs poorly compared to the WSARE algorithms. A closer analysis reveals that the difference in these results can be explained by the used evaluation measure. \citet{nss_fanaee} consider only $p$-values in the range $[0.02,0.25]$ to create the AMOC-curve. However, exactly the omitted low $p$-values are particularly important when precise predictions with low false positive rates are required which is why we explicitly included this range into the computation of the AMOC-curve.

\paragraph{Comparison to the Anomaly Detection Algorithms.}

Regarding the synthetic data, which was specifically created in order to evaluate the WSARE algorithms, we can observe that no anomaly detection algorithm can reach competitive $AAUC_{5\%}$ scores to WSARE 3.0. Considering the gap to WSARE 2.0, which in comparison to 3.0 does not distinguish between environmental settings, one reason could be that the anomaly detection algorithms are not able to take the environmental variables into account.
Another reason could be the low number of training instances (one for each day) which might have caused problems, especially for the neural networks. Only the SVM, which is known to work well with only few instances, and the Gaussian mixture model are able to achieve acceptable results. These two approaches are in fact able to outperform the WSARE variants on the real data for which we already found evidence that the environmental information might not be useful.

\paragraph{Comparison to the Benchmarks.} 
In the following, we will put the previously discussed results in relation to the benchmarks. For the global benchmarks, we can observe that monitoring the total number of cases per time slot is not sufficient to adequately detect most of the outbreaks. Notably, many of the machine learning approaches do in fact not perform considerably better than these simple benchmarks. 
The comparison to our proposed statistical benchmarks applied on each possible syndrome separately allow further important insights. Our main observation is that, despite its simplicity, they outperform most of the previously discussed, more sophisticated approaches. In fact, in the case of the real data the Gaussian and the negative binomial benchmarks achieve the best scores. On the synthetic data they are able to achieve results that are competitive to WSARE 3.0 even though the benchmarks do not take the environmental attributes into account.
We were also surprised by the good results of the Gaussian benchmark since this modelling is not specifically designed for count data. The advantage may be explained with the context of multiple testing and the generation of smoother, less extreme estimates and hence more reliable outlier scores for the time slots. However, the results on the real data, which obviously contain a more realistic representation of count data than the completely generated synthetic data, show that the negative binomial benchmark can improve over the Gaussian benchmark.

\paragraph{Comparison between $\mathcal{S}_{\leq1}$ and $\mathcal{S}_{\leq2}$.} 

We can make two basic observations regarding the complexity of the monitored syndromes: Firstly, the outbreaks in the synthetic data are better detected by the algorithms and benchmarks for non-specific syndromic surveillance when monitoring single condition syndromes $\mathcal{S}_{\leq1}$ while for the real data we benefit from pair patterns $\mathcal{S}_{\leq2}$. Secondly, almost no anomaly detector is able to profit from the explicit counts for $\mathcal{S}_{\leq2}$ regardless of the dataset. For understanding the first effect, we take a closer look at the results of our proposed benchmarks. These approaches can only take co-occurrences between conditions into account if explicitly given or if the $\mathcal{S} \setminus \mathcal{S}_{\leq1}$ patterns greatly affect the counts for the composing conditions. Hence, monitoring a larger set of syndromes increases the sensitivity of detecting outbreaks with complex disease patterns. However, it comes at the cost of a higher false alarm rate due to multiple testing. For the real dataset, for which we know that it contains more outbreaks based on two than on one condition, the higher sensitivity is able to outweigh the increased false alarm rate. On the other hand, the results on the synthetic dataset 
suggests that most of the outbreaks in the synthetic data are lead by single indicators, resulting in more false alarms when monitoring $\mathcal{S}_{\leq2}$.

In contrast to the non-specific syndromic surveillance approaches, only some anomaly detectors benefit and only slightly from the explicit counts for $\mathcal{S}_{\leq2}$, such as the local outlier factor algorithm and the isolation forests. This indicates that the remaining approaches, such as SVM and neural networks, already adequately consider correlations between attributes. 
Especially remarkable is the case of Gaussian mixture models, which achieves the best results in the group when monitoring $\mathcal{S}_{\leq1}$ but is strongly affected by the $\mathcal{S}_{\leq2}$ patterns.

\section{Conclusion}

In this work, we presented non-specific syndromic surveillance from the perspective of machine learning and gave an overview of the few approaches addressing this task. Furthermore, we introduced a way of how anomaly detection algorithms can be applied on this problem and a set of simple statistical algorithms which we believe should serve as reference points for future experimental comparisons. In an experimental evaluation, we revisited the non-specific syndromic surveillance approaches in face of the previously not considered statistical benchmarks and a variety of anomaly detectors. 
Eventually, we found that these benchmarks outperform most of the more sophisticated
techniques 
and are competitive to the best approaches in the field.


\remove{

For future work it remains, to improve the representation for the anomaly detectors or even include environmental variables


\todo{
- anomaly detection cannot differentiate between increases and decreases of syndrome counts. That is the advantage of our benchmarks. Future work, better representation for anomaly detection algorithms, maybe also considering environmental variables

}
}

%
%
%
\renewcommand{\bibsection}{\section*{\refname}} 
\footnotesize
\bibliographystyle{splncs04nat}
\bibliography{bib}

\end{document}